\documentclass{article}
\usepackage{algorithm}
\usepackage{bm}
\usepackage{appendix}
\usepackage{amsmath}
\usepackage{algpseudocode}
\usepackage{natbib}
\usepackage{amsthm}
\usepackage{graphicx}
\usepackage{multirow}
\usepackage{makebox}
\usepackage{booktabs}       
\usepackage{caption}
\usepackage{subfig}

     \usepackage[preprint]{neurips_2019}

\usepackage[utf8]{inputenc} 
\usepackage[T1]{fontenc}    
\usepackage{hyperref}       
\usepackage{url}            
\usepackage{booktabs}       
\usepackage{amsfonts}       
\usepackage{nicefrac}       
\usepackage{microtype}      
\newcommand\blfootnote[1]{%
  \begingroup
  \renewcommand\thefootnote{}\footnote{#1}%
  \addtocounter{footnote}{-1}%
  \endgroup
}

\title{Exploiting Uncertainty of Loss Landscape for Stochastic Optimization}

\author{Vineeth S. Bhaskara \\
Department of Computer Science\\
University of Toronto\\
\texttt{bhaskara@cs.toronto.edu} 
\And
Sneha Desai \\
Department of Computer Science\\
University of Toronto\\
\texttt{sdesai@cs.toronto.edu} 
}

\newcommand{\tta}{{\boldsymbol{\theta} } }   
   
\newcommand{\nablab}{{ \boldsymbol {{\nabla}} }}  
\newcommand{\nablabt}{{ \nablab_\tta }}

\begin{document}

\maketitle

\begin{abstract}
We introduce novel variants of momentum by incorporating the variance of the stochastic loss function. The variance characterizes the confidence or uncertainty of the local features of the averaged loss surface across the i.i.d. subsets of the training data defined by the mini-batches. We show two applications of the gradient of the variance of the loss function. First, as a bias to the conventional momentum update to encourage conformity of the local features of the loss function (e.g. local minima) across mini-batches to improve generalization and the cumulative training progress made per epoch. Second, as an alternative direction for "exploration" in the parameter space, especially, for non-convex objectives, that exploits both the optimistic and pessimistic views of the loss function in the face of uncertainty. We also introduce a novel data-driven stochastic regularization technique through the parameter update rule  that is model-agnostic and compatible with arbitrary architectures. We further establish connections to probability distributions over loss functions and the REINFORCE policy gradient update with baseline in RL. Finally, we incorporate the new variants of momentum proposed into Adam, and empirically show that our methods improve the rate of convergence of training based on our experiments on the MNIST and CIFAR-10 datasets.
\end{abstract}

\section{Introduction}

\blfootnote{Code for our optimizers and  experiments is publicly available at {\url{https://github.com/bsvineethiitg/adams}}.}Training deep neural networks by stochastic gradient descent has been highly successful in solving several important tasks in vision \citep{he2016deep}, language \citep{2019arXiv190410509C}, and Reinforcement Learning (RL) \citep{silver2017mastering}. Predominantly, the training procedure for modern deep neural networks involves some variation of vanilla stochastic gradient descent (SGD), where updates to the parameters are  based on the gradient computed over the current mini-batch's loss function.

The mini-batch gradient is a noisy unbiased estimator of the full-gradient. A widely used method to stabilize the mini-batch gradient is momentum, where parameter updates are based on an exponentially weighted average of the previous mini-batch gradients, thus "smoothing" out the oscillations in the updates \citep{goh2017momentum}. This improves training speed and convergence significantly \citep{sutskever2013importance}, and has remained an essential component of modern optimization algorithms such as Adam and AdaMax \citep{kingma2014adam}. In Section \ref{sec:momasfullg}, we present an alternative perspective on why momentum (with the bias-correction term) works by showing that it approximates the full-gradient under certain assumptions and an exponential probability distribution. 

In this paper, we propose exploiting the gradient of the second moment (or the "variance-gradient") of the stochastic loss function across mini-batches to quantify the uncertainty or error of the gradient of the first moment estimate (or the momentum) in approximating the full-gradient. The variance-gradient points along directions of the loss surface where the local features either conform or disagree the most across mini-batches.

In Section \ref{sec:momucb}, we introduce \textit{MomentumUCB}, a biased version of the momentum method that encourages updates along regions of the loss surface that locally conform across mini-batches in addition to the objective of minimizing the expected loss. In Section \ref{sec:momexplore}, we introduce \textit{MomentumCB} and \textit{MomentumS} that are biased and unbiased versions of the momentum, respectively, and exploit both the optimistic and pessimistic views of the loss surface in the face of uncertainty.

\section{Related work}

Previous  work such as SAG \citep{roux2012stochastic}, SAGA \citep{defazio2014saga} and SVRG \citep{johnson2013accelerating} propose accelerating SGD through variance reduction of the mini-batch gradient by introducing a baseline for the gradient that is computed every $m$ steps. 


Unlike the above work, our paper focuses on variance reduction of the underlying stochastic loss function rather than dealing directly with the variance of the gradient. Similar to SAG and SVRG, we introduce \textit{MomentumUCB}, a biased estimator that has an additional variance minimization objective. We also empirically show that instead of maximizing or minimizing the variance of the stochastic loss objective throughout the optimization, exploration in the parameter space by alternating between optimistic and pessimistic views of the loss landscape accelerates training and provides an unbiased estimate of the full-gradient. 

Similar to SAGA, we introduce \textit{MomentumS} that is an unbiased estimator. In contrast to SAG, SAGA and SVRG, our variants of momentum introduced in the paper are computationally similar in cost to SGD with conventional momentum.

\section{Momentum as an approximation to the full-gradient}
\label{sec:momasfullg}

\paragraph{Full-gradient} Consider the "full-loss" function $\mathcal{L}(\tta)$ under the parameters $\tta$ over the entire training dataset. Let the integer index $i\in [0, M-1]$ denote $i^\text{th}$ mini-batch out of a total of $M$ mini-batches of the training dataset. Then one may write the full-batch loss function as $\mathcal{L}(\tta) = \mathbb{E}_{\mathcal{P}(i)} \bigg[ \mathcal{L}^{(i)}(\tta) \bigg]$ under $\mathcal{P}(i)=\frac{1}{M}$. Therefore, the full-gradient at time step $t$ with parameters $\tta_{t-1}$, may be explicitly written as:
\begin{align}
\label{fullgrad}
\nablab_\tta \mathcal{L}(\tta_{t-1})  = \mathbb{E}_{\mathcal{P}(i)} \bigg[ \nablab_\tta\mathcal{L}^{(i)}(\tta_{t-1}) \bigg] =   \frac{1}{M} \bigg[  \nablabt \mathcal{L}^{(0)}(\tta_{t-1}) + \cdots + \nablabt \mathcal{L}^{(M-1)}(\tta_{t-1})    \bigg].
\end{align}

\paragraph{SGD with momentum and bias-correction term}
\begin{algorithm}[tp]
\caption{\textit{SGD with  Momentum and Bias-Correction term}.  All operations on vectors are element-wise. Default settings used are  $\beta=0.9$. Vectors are indicated in bold. }\label{alg:originalmom}
\begin{algorithmic}
\Require{$\alpha$: Stepsize}
\Require{$\beta \in[0,1)$: Exponential decay rate for momentum}
\Require{$ \mathcal{L}(\boldsymbol{\theta})$: Stochastic scalar loss objective function with parameters $\boldsymbol \theta$}
\Require{$\boldsymbol\theta_0$: Initial parameter vector}
\State $\boldsymbol m_0\gets \boldsymbol 0$ (Initialize $1^\text{{st}}$ moment vector of the mean loss gradient)
\State $ t\gets  0 $ (Initialize timestep)
\While{$\boldsymbol\theta_t$ not converged}
\State $ t\gets  t+1 $
\State $i \gets (t-1)~\verb|mod|(M)$ (Mini-batch index $i\in[0,M-1]$)
\State $\boldsymbol g_t\gets \boldsymbol \nabla_{\boldsymbol\theta}  \mathcal{L}^{(i)}(\tta_{t-1})$ (Get gradients w.r.t stochastic objective at timestep $t$ and mini-batch $i$)
\State $\boldsymbol m_t\gets \beta \cdot \boldsymbol m_{t-1} + (1-\beta)\cdot \boldsymbol g_t$ 
\State $ \widehat{\boldsymbol m}_t \gets \boldsymbol m_t/(1-\beta^t) $ (Bias-correction term)
\State $ {\boldsymbol \theta}_t \gets \boldsymbol \theta_{t-1} - \alpha \cdot {\widehat{\boldsymbol m}_t} $ (Update parameters)
\EndWhile
\State \textbf{return} ${\boldsymbol \theta}_t$ (Resulting parameters)
\end{algorithmic}
\end{algorithm}

Consider stochastic gradient descent with momentum and bias-correction term in Algorithm (\ref{alg:originalmom}). At time step $t$ (with the current mini-batch labeled by $i$), one may unroll the recurrence for $\boldsymbol m_t$ as 
\begin{align} 
\boldsymbol m_t &= (1-\beta)\cdot \boldsymbol g_t + (1-\beta)~\beta\cdot \boldsymbol g_{t-1} + (1-\beta)~\beta^2\cdot \boldsymbol g_{t-2} + \cdots + (1-\beta)~\beta^{t-1}\cdot \boldsymbol g_{1} \\ \label{momexpand}
&= (1-\beta)\cdot \nablabt \mathcal{L}^{(i)}(\tta_{t-1}) +  \cdots + (1-\beta)~\beta^{t-1}\cdot\nablabt \mathcal{L}^{(i-t+1)}(\tta_{0}). 
\end{align}
Note that the operations on index $i$ are all done in modulo $M$ so that the indices still represent one of the $M$ mini-batches. In Algorithm (\ref{alg:originalmom}), $\boldsymbol m_t$ corresponds to an \textit{exponentially weighted sum} (at timescale $\beta$) of the previous gradients across the mini-batches. The \textit{exponentially weighted average}, $ \widehat{\boldsymbol m}_t $, is obtained by dividing $\boldsymbol m_t$ by the sum of the exponential weights $ \sum_{j=0}^{j=t-1} (1-\beta)\beta^j$, which precisely gives the term $(1-\beta^t)$ that is referred to as the "bias-correction term" in Adam \citep{kingma2014adam}. Therefore, $ \widehat{\boldsymbol m}_t  = \boldsymbol m_t/(1-\beta^t)$ is the exponentially weighted average of the gradients across the mini-batches at time step $t$. Writing $ \widehat{\boldsymbol m}_t $ explicitly, we have
\begin{align}\label{momunroll}
\widehat{\boldsymbol m}_t =  \frac{1}{\{(1-\beta^t)/(1-\beta)\}} \bigg[  \nablabt \mathcal{L}^{(i)}(\tta_{t-1}) + \beta\cdot \nablabt \mathcal{L}^{(i-1)}(\tta_{t-2}) + \cdots + \beta^{t-1}\cdot\nablabt \mathcal{L}^{(i-t+1)}(\tta_{0}) \bigg].
\end{align}

Compare  Eq. \eqref{momunroll} with the  full-gradient in Eq. \eqref{fullgrad}. Since one may not computationally afford to evaluate the network at the current parameters $\tta_{t-1}$ for each mini-batch to get the  full-gradient at time step $t$, momentum  compromises to using an approximation for the full-gradient noting that $\tta_{t-1} \approx \tta_{t-2}$ when  the learning rate $\alpha$ is sufficiently small. An exponential decay weight proportional to $\beta^p$ is  considered for the gradient at parameters $\tta_{t-1-p}$ as the approximation \hbox{$\tta_{t-1} \approx \tta_{t-1-p}$} becomes less reasonable as $p$ gets larger. This can be noted by explicitly rewriting Eq. \eqref{momunroll} as an expectation under an  exponentially weighted  probability distribution $\mathcal{P}_{\beta}(p|t)$ at a given time step $t$ (defined below) as follows:
\begin{align}
    \label{momasexpectation}
    \widehat{\boldsymbol m}_t &= \mathbb{E}_{\mathcal{P}_{\beta}(p|t)}\left[ \nablabt \mathcal{L}^{(i)}(\tta_{t-1-p}) \Bigg| ~t,~p\in[0,~t-1],~i\equiv (t-1-p) ~\text{mod} (M) \right],~\text{where}\\
    \nonumber &\mathcal{P}_{\beta}(p|t) = \frac{\beta^{p}}{\{(1-\beta^t)/(1-\beta)\}}.~~(\text{Also note that}~\sum_{p=0}^{p=t-1}\mathcal{P}_{\beta}(p|t)=\frac{\sum_{p=0}^{p=t-1}\beta^{p}}{\{(1-\beta^t)/(1-\beta)\}}=1.)
\end{align}

Thus,  $\widehat{\boldsymbol m}_t$ can be viewed as an approximation to the  full-gradient $\nablabt ~{\mathcal{L}}(\tta_{t-1})$ at time step $t$, mini-batch index \hbox{$i=(t-1)~$\verb|mod|$(M)$}, and parameters $\tta_{t-1}$.

With this analogy in place, we approximate the gradient of the variance of the mini-batch loss function ("variance-gradient") under the exponentially weighted probability distribution defined above to derive an update rule similar to momentum.

\section{MomentumUCB: Biasing momentum along low variance regions of the loss landscape}
\label{sec:momucb}
Consider the loss objective that accounts the variance of the loss function to bias the updates along the regions of the landscape that conform across the mini-batches up to an extent determined by the "confidence hyperparameter" $\eta$ as follows:
\begin{align} \label{momucbloss} \mathcal{L}^{UCB}(\tta) =  \mathbb{E}_{i} \bigg[ \mathcal{L}^{(i)}(\tta) \bigg] + ~\eta \cdot \sqrt{\text{Var}_{i}\bigg[ \mathcal{L}^{(i)}(\tta) \bigg]},\end{align}
where the subscript $i$ corresponds to a probability distribution over the mini-batches and $\eta>0$. The above loss objective is similar in form to the Upper Confidence Bound (UCB) acquisition function in Bayesian optimization that is maximized to balance exploration with exploitation.

The additional variance-minimization objective encourages generalization by providing an incentive for performing equally well on individual i.i.d. subsets of the training data (defined by the mini-batches) separately to keep the variance of the loss lower.

\begin{figure}[tp]
\centering
\includegraphics[scale=0.5]{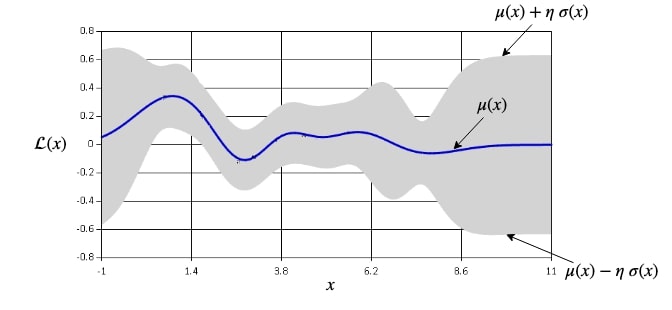}

  \caption{Using uncertainty to either choose an optimistic loss  surface that has a lower loss than the mean surface or a pessimistic loss surface that has a higher loss by $|\eta|$ standard deviations.}
  
  \label{fig:optimism_pessimism}
\end{figure}

Figure (\ref{fig:optimism_pessimism}) illustrates the resultant optimization landscape in 1D, as an example, for the cases of $\eta>0$ (pessimism) and $\eta>0$ (optimism) in the face of uncertainty.

Considering the gradient of $\mathcal{L}^{UCB}(\tta)$ in Eq. \eqref{momucbloss}, one has:
\begin{align}\label{momucbgrad}
\nablabt \mathcal{L}^{UCB}(\tta) 
&= \mathbb{E}_{i} \bigg[ \nablabt \mathcal{L}^{(i)}(\tta) \bigg] + \eta \cdot \nablabt ~ \sqrt{\text{Var}_{i}\bigg[ \mathcal{L}^{(i)}(\tta) \bigg]} \\
&= \mathbb{E}_{i} \bigg[ \nablabt \mathcal{L}^{(i)}(\tta) \bigg] +   \frac{\eta}{\sigma_l} \cdot \left\{   \mathbb{E}_{i} \bigg[ \mathcal{L}^{(i)}(\tta)\cdot\nablabt \mathcal{L}^{(i)}(\tta) \bigg] - \mu_l\cdot   \mathbb{E}_{i} \bigg[ \nablabt \mathcal{L}^{(i)}(\tta) \bigg] \right\}\\
&= \mathbb{E}_{i} \bigg[ \left( 1 + \eta ~~ \frac{\mathcal{L}^{(i)}(\tta) - \mu_l}{\sigma_l} \right) \cdot  \nablabt \mathcal{L}^{(i)}(\tta)       \bigg],          \label{expandedmomucb}
\end{align}
where $\sigma_l^2=\text{Var}_{i} [\mathcal{L}^{(i)}(\tta)] = \mathbb{E}_{i} \left[  \left(\mathcal{L}^{(i)}(\tta)\right)^2 \right] -  \left(\mathbb{E}_{i} \big[ \mathcal{L}^{(i)}(\tta) \big]\right)^2$, and $\mu_l=\mathbb{E}_{i} \bigg[ \mathcal{L}^{(i)}(\tta) \bigg]$.

We infer the update rule for computing $\nablabt \mathcal{L}^{UCB}(\tta)$ by approximating the  expectation  in Eq. \eqref{expandedmomucb} under the exponentially weighted probability distribution $\mathcal{P}_\beta(p|t)$. We refer to the resultant update rule as  \textit{MomentumUCB}  to distinguish it from the conventional momentum update.

\paragraph{\textit{AdamUCB}: Adam with MomentumUCB} Considering the stochastic gradient at time step $t$ to be \hbox{$\boldsymbol g_t=\nablabt\mathcal{L}^{(i)}(\tta_{t-1})$}, one may write the traditional Adam update concisely as $\delta\tta_{\text{Adam}} = \frac{\langle \boldsymbol g_t \rangle_{\beta_1}}{\sqrt{\langle \boldsymbol g_t^2 \rangle_{\beta_2}}}$, where $\beta$s are the time scales, and $\langle\cdot\rangle_\beta$ denotes  $\mathbb{E}_{\mathcal{P}_\beta}[\cdot]$ under the exponentially decaying probability distribution. \label{adam1}

We introduce \textit{AdamUCB} in Algorithm (\ref{alg:adamucb}) that implements \textit{MomentumUCB} (Eq. \eqref{expandedmomucb}) in the parameter update rule as follows
\begin{align} \delta\tta_{\text{AdamUCB}} &= \frac{\left\langle  \left( 1+\eta~\frac{l_t-\mu_l}{\sigma_l}\right) \boldsymbol g_t \right\rangle_{\beta_1}}{\sqrt{\left\langle \left( 1+\eta~\frac{l_t-\mu_l}{\sigma_l}\right)^2 \boldsymbol g_t^2 \right\rangle_{\beta_2}}}=\frac{\Big\langle  \left[ \sigma_l +\eta\cdot(l_t-\mu_l)\right] \boldsymbol g_t \Big\rangle_{\beta_1}}{\sqrt{\left\langle \left[ \sigma_l +\eta\cdot(l_t-\mu_l)\right]^2 \boldsymbol g_t^2 \right\rangle_{\beta_2}}},\label{adamucbrule}\end{align}
where $l_t=\mathcal{L}^{(i)}(\tta_{t-1})$ is the  mini-batch loss  at the current time step $t$, \{$\mu_l,~\sigma_l$\} are the mean and the standard deviation of  mini-batch losses up to time step $t-1$, respectively, and $\eta$ is the confidence hyperparameter.

\begin{algorithm}
\caption{\textit{AdamUCB: Upper Confidence Bounded Adam}, our proposed algorithm for stochastic optimization. All operations on vectors are element-wise. $\eta$ is the confidence hyperparameter. Defaults for $\beta_1$, $\beta_2$ and $\epsilon$ are 0.9, 0.999, $10^{-8}$ respectively (same as Adam). Vectors are indicated in bold. (Using $\eta=0$ restores the original Adam update.)}\label{alg:adamucb}
\begin{algorithmic}
\Require{$\alpha$: Stepsize}
\Require{$\beta_1,~\beta_2 \in[0,1)$: Exponential decay rates for moment estimates}
\Require{$\eta$: Exploration or the upper confidence parameter ($\eta>0$) }
\Require{$\mathcal{L}(\boldsymbol{\theta})$: Stochastic scalar loss objective function with parameters $\boldsymbol \theta$}
\Require{$\boldsymbol\theta_0$: Initial parameter vector}
\State $\boldsymbol m_0\gets \boldsymbol 0$ (Initialize UCB weighted $1^\text{{st}}$ moment gradient vector)
\State $\boldsymbol v_0\gets \boldsymbol 0$ (Initialize UCB weighted $2^\text{{nd}}$ moment gradient vector)
\State $ r_0\gets  0$ (Initialize $1^\text{{st}}$ moment of loss)
\State $ s_0\gets  0$ (Initialize $2^\text{{nd}}$ moment of loss)
\State $ t\gets  0 $ (Initialize timestep)
\While{$\boldsymbol\theta_t$ not converged}
\State $ t\gets  t+1 $
\State $i \gets (t-1)~\verb|mod|(M)$ (Mini-batch index $i\in[0,M-1]$)
\State $\boldsymbol g_t\gets \boldsymbol \nabla_{\boldsymbol\theta}  \mathcal{L}^{(i)}(\tta_{t-1})$ (Get gradients w.r.t stochastic objective at timestep $t$ and mini-batch $i$)
\State $ l_t\gets \mathcal{L}^{(i)}(\tta_{t-1})$ (Get the stochastic objective evaluated at timestep $t$ and mini-batch $i$)
\State $\mu_l \gets r_{t-1}/(1-\beta_1^{t-1}+\epsilon)$ (Average loss till $t-1$ timestep)
\State $\widehat{s}_{t-1} \gets s_{t-1}/(1-\beta_1^{t-1}+\epsilon)$ (Average squared loss till $t-1$ timestep)
\State $\sigma_l \gets \sqrt{\widehat{s}_{t-1} - \mu_l^2}$ (Standard deviation of loss till $t-1$ timestep)
\State $\boldsymbol m_t\gets \beta_1 \cdot \boldsymbol m_{t-1} + (1-\beta_1)\cdot [\sigma_l + \eta~(l_t - \mu_l) ] \cdot \boldsymbol g_t$ 
\State $\boldsymbol v_t\gets \beta_2 \cdot \boldsymbol v_{t-1} + (1-\beta_2)\cdot [\sigma_l + \eta~(l_t - \mu_l) ]^2 \cdot \boldsymbol g_t^2$ 
\State $ r_t\gets \beta_1 \cdot  r_{t-1} + (1-\beta_1)\cdot  l_t$
\State $ s_t\gets \beta_1 \cdot  s_{t-1} + (1-\beta_1)\cdot  l_t^2$
\State $ \widehat{\boldsymbol m}_t \gets \boldsymbol m_t/(1-\beta_1^t) $
\State $ \widehat{\boldsymbol v}_t \gets \boldsymbol v_t/(1-\beta_2^t) $

\State $ {\boldsymbol \theta}_t \gets \boldsymbol \theta_{t-1} - \alpha \cdot \widehat{\boldsymbol m}_t/\left({\sqrt{\widehat{\boldsymbol v}_t }+ \epsilon}\right) $ (Update parameters)
\EndWhile
\State \textbf{return} ${\boldsymbol \theta}_t$ (Resulting parameters)
\end{algorithmic}
\end{algorithm}

\subsection{Connections to policy gradient with baseline in reinforcement learning}
\label{sec:policygradRL}
Consider a typical setting of a RL problem with  $r(\tau)$ representing the cumulative reward for a roll out  $\tau$. 
The goal is to maximize the expected return, \hbox{$R=\mathbb{E}_{P(\tau)}[r(\tau)]$}, where $P(\tau)$ represents the probability of the roll out $\tau$ that depends both on the policy and the dynamics of the environment.

The REINFORCE policy gradient update with baseline $b$ can be written as 
\hbox{$\nablabt \mathbb{E}_{P(\tau)}[r(\tau)] = \mathbb{E}_{P(\tau)}\left[(r(\tau)-b)\cdot\nablabt\log P(\tau)\right]$}. A common choice for the baseline is the average return obtained so far, i.e., $b=\mathbb{E}_{P(\tau)}[r(\tau)]$. Substituting into REINFORCE, we have
\begin{align}
    \nablabt \mathbb{E}_{P(\tau)}[r(\tau)]&=\mathbb{E}_{P(\tau)}\left[(r(\tau)-\mathbb{E}_{P(\tau)}[r(\tau)])\cdot\nablabt\log P(\tau)\right]\\
    &=\mathbb{E}_{P(\tau)}\left[r(\tau)\cdot\nablabt\log P(\tau)\right] - \mathbb{E}_{P(\tau)}[r(\tau)] \cdot \mathbb{E}_{P(\tau)}\left[\nablabt\log P(\tau)\right]\\
    &=\text{Cov}_{P(\tau)}\left[ r(\tau),  \nablabt\log P(\tau) \right].\label{CovRL}
\end{align}

Consider the MomentumUCB gradient $\nablabt \mathcal{L}^{UCB}(\tta)$ from Eq. \eqref{expandedmomucb} as follows:
\begin{align}
    \mathbb{E}_{i} \bigg[ \left( 1 + \eta  \frac{\mathcal{L}^{(i)}(\tta) - \mu_l}{\sigma_l} \right) \cdot  \nablabt \mathcal{L}^{(i)}(\tta)       \bigg]=\underbrace{\mathbb{E}_{i} \bigg[ \nablabt \mathcal{L}^{(i)}(\tta)\bigg]}_{\text{conventional momentum}} + \frac{\eta}{\sigma_l}\cdot  \underbrace{\mathbb{E}_{i} \bigg[ \left(\mathcal{L}^{(i)}(\tta) - \mu_l\right)\cdot\nablabt \mathcal{L}^{(i)}(\tta)\bigg]}_{\text{variance-gradient}}.\label{covtermucb}
\end{align}
Simplifying the variance-gradient term, we have
\begin{align} 
\mathbb{E}_{i} \bigg[ \left(\mathcal{L}^{(i)}(\tta) - \mu_l\right)\cdot\nablabt \mathcal{L}^{(i)}(\tta)\bigg]
&= \mathbb{E}_{i} \bigg[ \mathcal{L}^{(i)}(\tta) \cdot\nablabt \mathcal{L}^{(i)}(\tta)\bigg] - \mathbb{E}_{i} \bigg[  \mathcal{L}^{(i)}(\tta)\bigg] \cdot \mathbb{E}_{i} \bigg[ \nablabt \mathcal{L}^{(i)}(\tta)\bigg]\nonumber\\
&=\text{Cov}_{i}\left[ \mathcal{L}^{(i)}(\tta),~\nablabt\mathcal{L}^{(i)}(\tta) \right].\label{momucbrl}
\end{align}

Therefore, when $r(\tau)$ and $\log P(\tau)$ in Eq. \eqref{CovRL} are chosen to be the negative cross-entropy loss $-L_{CE}=\log P(\text{target}|\text{data})$ in a supervised learning setting, for instance, and the co-variance is instead computed across the mini-batches, then the policy gradient term in Eq. \eqref{CovRL} reduces to the  variance-gradient in \hbox{Eq. \eqref{momucbrl}}.

\section{Exploiting pessimism and optimism in the face of uncertainty}
\label{sec:momexplore}
In the previous section, we introduced MomentumUCB justifying the case for $\eta>0$ as taking a pessimistic view of the loss surface that biases updates along regions that conform across  mini-batches. An undesirable effect of such a variance-minimization objective is that the parameters might eventually land on a plateau of the loss surface, preventing further progress and slowing down training. In this section we propose  incorporating the best of both the cases of $\eta>0$ (pessimism in the face of uncertainty) and $\eta<0$ (optimism in the face of uncertainty) where the updates alternate between maximizing and minimizing the variance objective based on a criterion.

\subsection{Bounding the relative standard deviation by  $\eta$ on both sides}

We propose a simple modification to the MomentumUCB term in Eq. \eqref{expandedmomucb} by replacing $\eta$ with the difference of the current relative standard deviation (defined by $\frac{\sigma_l}{|\mu_l|}$) and the required relative standard deviation (specified by a hyperparameter $\eta$ that is \textit{different} from the $\eta$ in MomentumUCB) that must be maintained throughout the optimization. The intuition behind this criterion is that the variance of the loss landscape should not be too low (to avoid undesirable plateaus of the surface) or too high (since a "good" minima likely performs comparably well across the i.i.d. mini-batches).

Replacing $\eta \rightarrow -\left( \eta - \frac{\sigma_l}{|\mu_l|}\right)$, we have the following version of Momentum that we call \textit{MomentumCB} or \textit{Confidence bounded Momentum} since the standard deviation is bounded on both the sides (two-sided bound):
\begin{align}
    \underbrace{\frac{1}{\sigma_l}~\Big\langle  \left[ \sigma_l +\eta\cdot(l_t-\mu_l)\right] \boldsymbol g_t \Big\rangle}_{\text{MomentumUCB (one-sided bound)}} ~\longrightarrow ~ \underbrace{\frac{1}{\sigma_l}~\Big\langle  \left[ \sigma_l -\left( \eta - \frac{\sigma_l}{|\mu_l|}\right)\cdot(l_t-\mu_l)\right] \boldsymbol g_t \Big\rangle}_{\text{MomentumCB (two-sided bound)}}.
\end{align}
When the current relative std. dev.  is greater than the specified hyperparameter $\eta$, the term $\left[-\left( \eta - \frac{\sigma_l}{|\mu_l|}\right)\right]$ takes a positive sign (with a magnitude proportional to the violation) and  updates along directions that reduce the variance (in addition to the usual momentum gradient), and, hence, takes a pessimistic view of the loss surface. Similarly, when the current relative std. dev. is lower than the hyperparameter $\eta$, the updates get biased along directions that increase the variance.

\paragraph{\textit{AdamCB}: Adam with MomentumCB} We incorporate MomentumCB into Adam under the exponential probability distribution (similar to Algorithm (\ref{alg:adamucb})). The parameter update rule for \textit{AdamCB} is given by
\begin{align} 
 \delta\tta_{\text{AdamCB}} =\label{adam2}\frac{\Big\langle  \left[ \sigma_l -\left( \eta - \frac{\sigma_l}{|\mu_l|}\right)(l_t-\mu_l)\right] \boldsymbol g_t \Big\rangle_{\beta_1}}{\sqrt{\Big\langle  \left[ \sigma_l -\left( \eta - \frac{\sigma_l}{|\mu_l|}\right)(l_t-\mu_l)\right]^2 \boldsymbol g_t^2 \Big\rangle_{\beta_2}}}=\frac{\Big\langle  \left[ \sigma_l ~ |\mu_l| -\left( \eta  ~|\mu_l| - \sigma_l\right)(l_t-\mu_l)\right] \boldsymbol g_t \Big\rangle_{\beta_1}}{\sqrt{\Big\langle  \left[ \sigma_l ~ |\mu_l| -\left( \eta ~ |\mu_l| - \sigma_l\right)(l_t-\mu_l)\right]^2 \boldsymbol g_t^2 \Big\rangle_{\beta_2}}}.\nonumber
\end{align}

\subsection{The reparametrization trick and stochastic momentum}
\label{stochamom}
In this section we introduce a stochastic regularizer based on the variance-gradient  that utilizes both the directions of minimizing and maximizing the variance to randomly "explore" in the parameter space. 

By "exploration," especially, in the context of non-convex optimization, we refer to choosing a perturbed parameter based on the variance-gradient after a conventional momentum update step. This acts as an initialization for the subsequent update and allows access to multiple regions of the loss surface over the course of training. Also, unlike dropout \citep{srivastava2014dropout}, our stochastic regularizer is architecture and model agnostic, and, therefore, is compatible with batch normalization \citep{2018arXiv180105134L}.

If $\eta$ in the  Eq. \eqref{momucbloss} is promoted to a new random variable $\hat{N}$ such that $\hat{N} \sim \mathcal{N}(0,~\eta)$,
where $\mathcal{N}(\cdot)$ is a Gaussian distribution and $\eta$ specifies the variance of the Gaussian (different from $\eta$ in MomentumUCB), then we have the  loss function $\mathcal{L}^{UCB}(\tta)$  also promoted to a random variable $\hat{\mathcal{L}}(\tta)$ in $\hat{N}$ such that 
\[ \hat{\mathcal{L}}(\tta) =  \mathbb{E}_{i} \bigg[ \mathcal{L}^{(i)}(\tta) \bigg] + \hat{N} \cdot \sqrt{\text{Var}_{i}\bigg[ \mathcal{L}^{(i)}(\tta) \bigg]}.\]~

Therefore, instead of fixing $\eta$ in MomentumUCB, if one samples it from a Gaussian distribution centered around zero with a specified standard deviation (given by a new hyperparameter $\eta$), then one can exploit both the directions of minimizing and maximizing the variance during the optimization. 

By the reparametrization trick \citep{kingma2013auto}, sampling $\hat{N} \sim \mathcal{N}(0,~\eta)$ in the above equation for $\hat{\mathcal{L}}$ is equivalent to sampling a loss function from a Gaussian distribution over mini-batch loss functions, i.e., 
\[\text{when}~\hat{N} \sim \mathcal{N}(0,~\eta) \implies \hat{\mathcal{L}} \sim \mathcal{N}\left(\mathbb{E}_{i} \bigg[ \mathcal{L}^{(i)}(\tta) \bigg], ~\eta\cdot\sqrt{\text{Var}_{i}\bigg[ \mathcal{L}^{(i)}(\tta) \bigg]}\right), \]
before computing the gradient. We call the approximation of $\nablabt\hat{\mathcal{L}}(\tta)$ under the exponential probability distribution $\mathcal{P}_\beta(p|t)$ as  \textit{Stochastic Momentum} or \textit{MomentumS} since $\hat{\mathcal{L}}(\tta)$ is a stochastic \hbox{variable in $\hat{N}$}.

Interestingly, since $\hat{\mathcal{L}}$ is  a random variable in $\hat{N}$, its expectation over the Gaussian $\mathcal{N}(\cdot)$ results in an unbiased estimate of the full-gradient. That is, 
\[ \mathbb{E}_{\mathcal{N}}[\hat{\mathcal{L}}] =  \mathbb{E}_{i} \bigg[ \mathcal{L}^{(i)}(\tta) \bigg] + \mathbb{E}_{\mathcal{N}}[\hat{N}] \cdot \sqrt{\text{Var}_{i}\bigg[ \mathcal{L}^{(i)}(\tta) \bigg]} = \mathbb{E}_{i} \bigg[ \mathcal{L}^{(i)}(\tta) \bigg], \]
since $\mathbb{E}_{\mathcal{N}}[\hat{N}]=0$ when $\mathcal{N}\equiv\mathcal{N}(0,~\eta)$. Therefore, \textit{MomentumS} is an unbiased estimate of the full-gradient when  $\hat{N}$ is sampled from 
$\mathcal{N}(0,~\eta)$ at each step.

With recent  high-capacity deep neural networks such as OpenAI Five \citep{OpenAI_dota} being trained  for several months continuously, "exploration" in the parameter space ensures that the optimization is not wastefully stuck on a plateau or bounded within a local region.

\paragraph{\textit{AdamS}: Adam with Stochastic Momentum} We incorporate  \textit{Stochastic Momentum}  into Adam by sampling $\eta$ in AdamUCB (see Algorithm (\ref{alg:adamucb})) from a zero-centered Gaussian whose standard deviation is provided as a hyperparameter.    \label{adam3}

\section{Experiments}
\label{experiments}
\begin{figure}[tp]
\centering
\makebox[\textwidth][c]{
 \subfloat[Training loss]{ \includegraphics[scale=0.35]{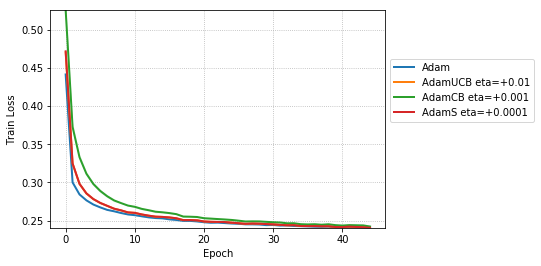}}
 \subfloat[Validation loss]{\includegraphics[scale=0.35]{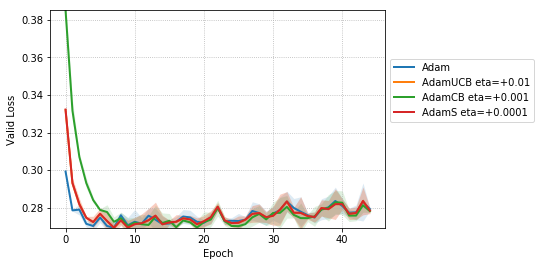}}
 \subfloat[Validation acc.]{\includegraphics[scale=0.35]{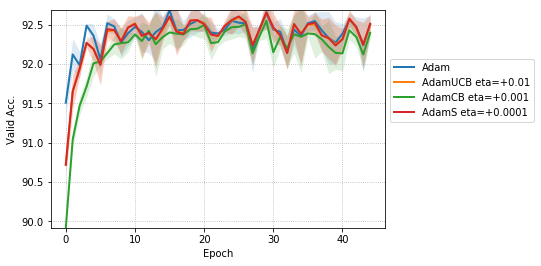}}
  
 } 

   \caption{\label{figmain:bestmnistlr} Comparison of optimizers for the LR model trained on MNIST that has a convex objective. No data augmentation was used.}
   
\end{figure}

\begin{figure}[tp]
\centering
\makebox[\textwidth][c]{
 \subfloat[Train loss]{\includegraphics[scale=0.35]{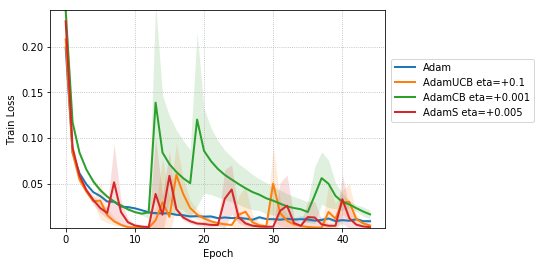}}
 \subfloat[Validation loss]{\includegraphics[scale=0.35]{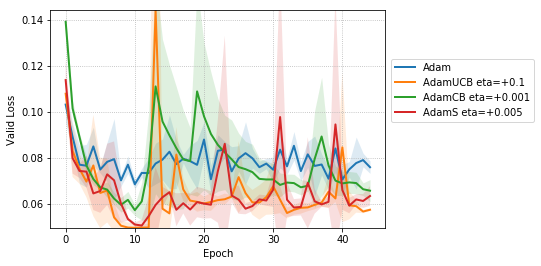}}
 \subfloat[Validation acc.]{\includegraphics[scale=0.35]{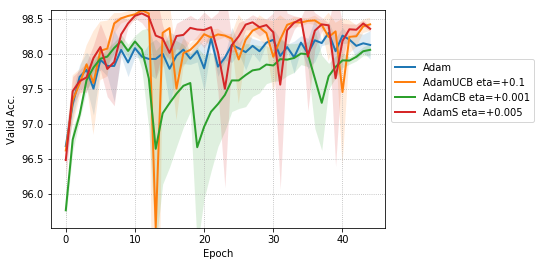}}
 
 }

   \caption{\label{figmain:bestetasmnistmlp} Comparison of optimizers  for a 2-hidden layer  MLP model with 1000 ReLU units each trained on MNIST. No data augmentations or dropout was used.}
   

\end{figure}

\begin{figure}[tp]
\centering
\makebox[\textwidth][c]{
 \subfloat[Training/Validation loss and Validation accuracy when the CNN is trained without dropout at \hbox{batch size $=128$}.]{ \includegraphics[scale=0.35]{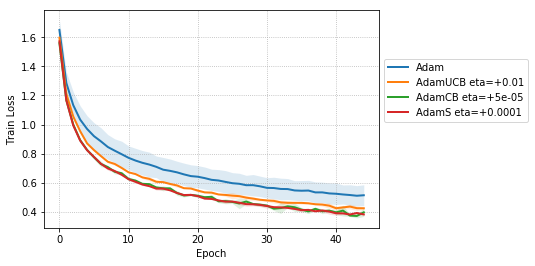}
 \includegraphics[scale=0.35]{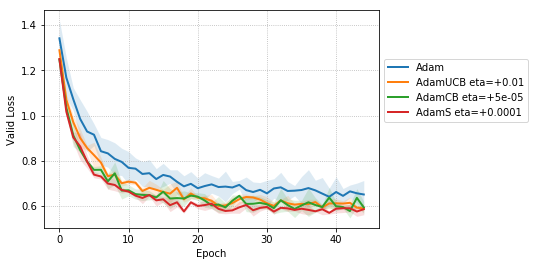}
 \includegraphics[scale=0.35]{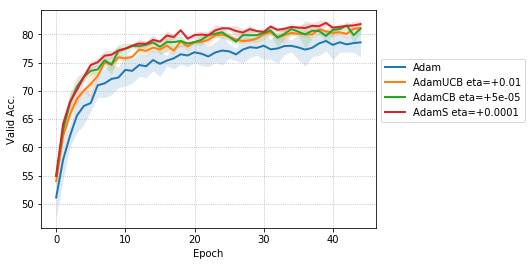}
 } 
 } \\
 
 \makebox[\textwidth][c]{
 \subfloat[Training/Validation loss and Validation accuracy when the CNN is trained with dropout at \hbox{batch size $=128$}.]{ \includegraphics[scale=0.35]{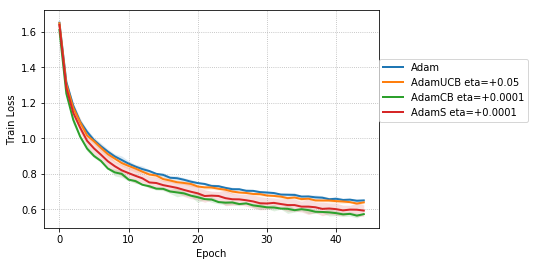}
 \includegraphics[scale=0.35]{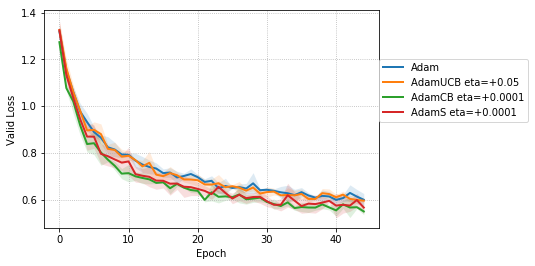}
 \includegraphics[scale=0.35]{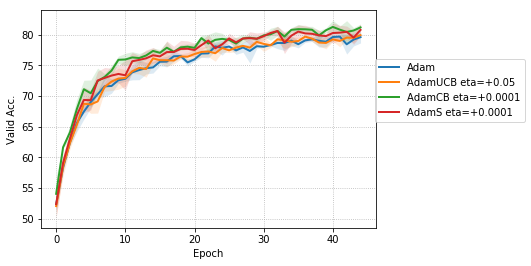}
 } }

   \caption{\label{figmain:bestetasperformancecifar10} Comparison of optimizers on a CNN model with c64-c64-c128-1000 architecture trained on CIFAR-10 (a) without dropout, and (b) with dropout ($p_{drop}=0.5$) on the inputs to the FC-1000 layer. Each conv layer of size $5\times5$  (stride=1, padding=2) is followed by a ReLU non-linearity and a max pooling layer of size $3\times3$ (stride=2). Random crops (with 4 padded pixels) and horizontal flips were considered for augmenting the dataset.}
   

\end{figure}

We empirically evaluate the three variants of Adam, namely, AdamUCB, AdamCB and AdamS  and compare their performance with the original Adam optimizer. We train multiple architectures of neural networks such as logistic regression (see Figure \ref{figmain:bestmnistlr}), MLPs (see Figure \ref{figmain:bestetasmnistmlp}), CNNs (see Figure \ref{figmain:bestetasperformancecifar10}) on MNIST/CIFAR-10 datasets on a single nVIDIA Tesla P4 GPU. The architectures of the networks are chosen to closely resemble the experiments published by \cite{kingma2014adam}.

Since our comparison is only among Adam-like optimizers, we use a fixed learning rate of $\alpha=0.001$ (without any scheduling) and do not search over different $\alpha$s. This is because the RMSprop-like denominator in all of our proposed variants ensures the step size to be roughly the same as Adam. We also use a $L2$ weight-decay of $10^{-4}$, batch size of 128, and keep the values of $\beta_1$ and $\beta_2$ fixed to Adam defaults of 0.9 and 0.999, respectively, across all our experiments. The input images are pre-processed by normalizing with the mean and the standard deviation of the pixel values. Best hyperparameter $\eta$ obtained  by searching  over a grid is used for each variant of the optimizer in the comparison. We present detailed results for additional training configurations (such as different batch sizes, etc) in the appendices.

\section{Discussion}

For the simple case of logistic regression on MNIST, the original Adam algorithm performs the best across the training metrics (Figure \ref{figmain:bestmnistlr}). Since convex objectives have an unique solution, the advantage of "exploration" along variance-gradient direction diminishes. 

From Figure \ref{figmain:bestetasmnistmlp}, for the case of MLPs trained on MNIST, AdamS and AdamUCB achieve a lower training and validation loss on average than Adam. For instance, at epoch 20 and 45, AdamS achieves half and one-third of the training error of Adam, respectively.

For CNNs  trained on CIFAR-10 (Figure \ref{figmain:bestetasperformancecifar10}), AdamUCB, AdamCB and AdamS perform significantly better the original Adam optimizer when no dropout is used ($\approx$ \textbf{6\%} improvement in val. acc. at epoch = 3). Not only is the validation loss better for our variants of Adam but also is the rate of convergence of training. We notice that AdamCB and AdamS consistently perform better than AdamUCB in this case. This shows that exploiting both the directions of variance-gradient indeed helps. For the case of CNNs with dropout, AdamCB and AdamS still outperform Adam but with a reduced margin of improvement ($\approx$  \textbf{2\%} improvement in val. acc. at epoch = 3).

\begin{figure}[tp]
\centering
\makebox[\textwidth][c]{
 \subfloat[Training loss]{ \includegraphics[scale=0.33]{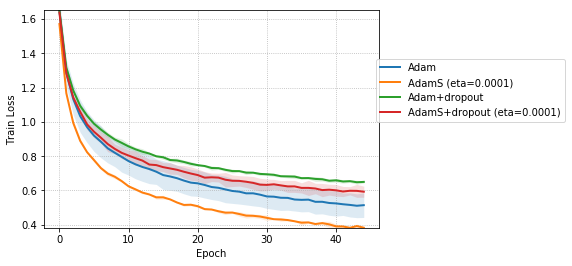}}
 \subfloat[Validation loss]{\includegraphics[scale=0.33]{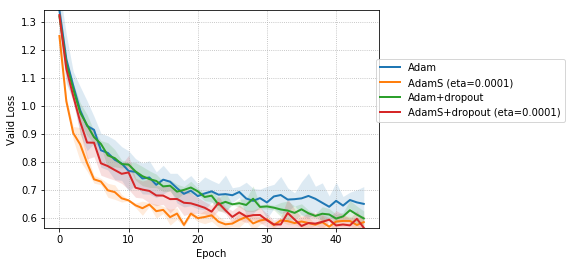}}
 \subfloat[Validation acc.]{\includegraphics[scale=0.33]{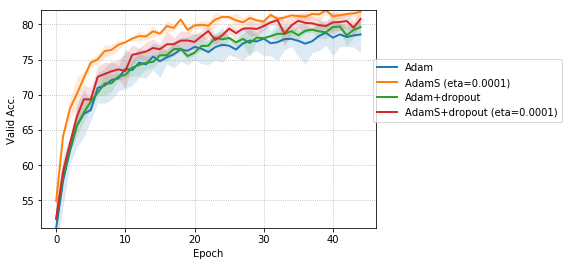}}
  
 } 

   \caption{\label{figmain:dropoutadams} Comparison of Dropout with the stochastic regularization implemented by AdamS for the CNN model c64-c64-c128-1000 trained on CIFAR-10 dataset.}
   
\end{figure}

Figure \ref{figmain:dropoutadams}  compares the effect of adding dropout to Adam and AdamS for the case of CNNs trained on CIFAR-10 dataset. The stochastic regularization implemented by AdamS leads to faster training convergence and better validation error when compared to dropout regularization.

\section{Conclusion and future work}

In this paper, we introduced novel ways of incorporating the variance information of the loss landscape across mini-batches for stochastic optimization. Based on our experiments with CIFAR-10, we recommend AdamS for optimizing general non-convex objectives (a good default for $\eta$ is $0.0001$).

Our work opens up directions of incorporating existing research on exploration--exploitation trade-off in Bayesian optimization into gradient-based stochastic optimization algorithms. Interesting directions for future research include exploiting other acquisition functions like Probability of Improvement (PI), Expected Improvement (EI), among others, to design loss objectives that efficiently utilize the uncertainty information of the loss landscape to accelerate training.

Investigating  SGD  with variations of momentum proposed  in this paper could also prove to be interesting as our formulation naturally gives a schedule for the learning rate through the variance of the loss. Finally, analyzing the effect of decay schedules for $\eta$, AdaMax-like modifications to the proposed variants of Adam, and incorporating Nesterov's momentum-like update rule \citep{nesterov1983method, dozat2016incorporating} would be other interesting directions to pursue. 

\subsection*{Contributions}
\textit{V.S.B.} contributed to the theory, derivations, and  the experiments on  CIFAR-10 dataset using CNN architectures. \textit{S.D.} verified the derivations and contributed to the experiments on MNIST using Logistic Regression and MLPs.

\subsubsection*{Acknowledgments}
We greatly acknowledge the supervision of our project by Prof. Jimmy Ba and Prof. Roger Grosse at the University of Toronto. We also acknowledge the Vector Institute Scholarship in Artificial Intelligence (VSAI) for supporting our graduate studies towards MSc in Applied Computing (MScAC).  


\begin{thebibliography}{15}
\providecommand{\natexlab}[1]{#1}
\providecommand{\url}[1]{\texttt{#1}}
\expandafter\ifx\csname urlstyle\endcsname\relax
  \providecommand{\doi}[1]{doi: #1}\else
  \providecommand{\doi}{doi: \begingroup \urlstyle{rm}\Url}\fi

\bibitem[{Child} et~al.(2019){Child}, {Gray}, {Radford}, and
  {Sutskever}]{2019arXiv190410509C}
Rewon {Child}, Scott {Gray}, Alec {Radford}, and Ilya {Sutskever}.
\newblock {Generating long sequences with sparse transformers}.
\newblock \emph{arXiv e-prints}, art. arXiv:1904.10509, Apr 2019.

\bibitem[Defazio et~al.(2014)Defazio, Bach, and
  Lacoste-Julien]{defazio2014saga}
Aaron Defazio, Francis Bach, and Simon Lacoste-Julien.
\newblock Saga: A fast incremental gradient method with support for
  non-strongly convex composite objectives.
\newblock In \emph{Advances in Neural Information Processing Systems}, pp.\
  1646--1654, 2014.

\bibitem[Dozat(2016)]{dozat2016incorporating}
Timothy Dozat.
\newblock Incorporating nesterov momentum into adam.
\newblock 2016.

\bibitem[Goh(2017)]{goh2017momentum}
Gabriel Goh.
\newblock Why momentum really works.
\newblock \emph{Distill}, 2\penalty0 (4):\penalty0 e6, 2017.

\bibitem[He et~al.(2016)He, Zhang, Ren, and Sun]{he2016deep}
Kaiming He, Xiangyu Zhang, Shaoqing Ren, and Jian Sun.
\newblock Deep residual learning for image recognition.
\newblock In \emph{Proceedings of the IEEE Conference on Computer Vision and
  Pattern Recognition}, pp.\  770--778, 2016.

\bibitem[Johnson \& Zhang(2013)Johnson and Zhang]{johnson2013accelerating}
Rie Johnson and Tong Zhang.
\newblock Accelerating stochastic gradient descent using predictive variance
  reduction.
\newblock In \emph{Advances in Neural Information Processing Systems}, pp.\
  315--323, 2013.

\bibitem[Kingma \& Ba(2014)Kingma and Ba]{kingma2014adam}
Diederik~P Kingma and Jimmy Ba.
\newblock Adam: A method for stochastic optimization.
\newblock \emph{arXiv preprint arXiv:1412.6980}, 2014.

\bibitem[Kingma \& Welling(2013)Kingma and Welling]{kingma2013auto}
Diederik~P Kingma and Max Welling.
\newblock Auto-encoding variational bayes.
\newblock \emph{arXiv preprint arXiv:1312.6114}, 2013.

\bibitem[{Li} et~al.(2018){Li}, {Chen}, {Hu}, and {Yang}]{2018arXiv180105134L}
Xiang {Li}, Shuo {Chen}, Xiaolin {Hu}, and Jian {Yang}.
\newblock {Understanding the disharmony between dropout and batch normalization
  by variance shift}.
\newblock \emph{arXiv e-prints}, art. arXiv:1801.05134, Jan 2018.

\bibitem[Nesterov(1983)]{nesterov1983method}
Yurii~E Nesterov.
\newblock A method for solving the convex programming problem with convergence
  rate o (1/k\^{} 2).
\newblock In \emph{Dokl. akad. nauk Sssr}, volume 269, pp.\  543--547, 1983.

\bibitem[OpenAI(2018)]{OpenAI_dota}
OpenAI.
\newblock Openai five.
\newblock \url{https://blog.openai.com/openai-five/}, 2018.

\bibitem[Roux et~al.(2012)Roux, Schmidt, and Bach]{roux2012stochastic}
Nicolas~L Roux, Mark Schmidt, and Francis~R Bach.
\newblock A stochastic gradient method with an exponential convergence \_rate
  for finite training sets.
\newblock In \emph{Advances in Neural Information Processing Systems}, pp.\
  2663--2671, 2012.

\bibitem[Silver et~al.(2017)Silver, Schrittwieser, Simonyan, Antonoglou, Huang,
  Guez, Hubert, Baker, Lai, Bolton, et~al.]{silver2017mastering}
David Silver, Julian Schrittwieser, Karen Simonyan, Ioannis Antonoglou, Aja
  Huang, Arthur Guez, Thomas Hubert, Lucas Baker, Matthew Lai, Adrian Bolton,
  et~al.
\newblock Mastering the game of go without human knowledge.
\newblock \emph{Nature}, 550\penalty0 (7676):\penalty0 354, 2017.

\bibitem[Srivastava et~al.(2014)Srivastava, Hinton, Krizhevsky, Sutskever, and
  Salakhutdinov]{srivastava2014dropout}
Nitish Srivastava, Geoffrey Hinton, Alex Krizhevsky, Ilya Sutskever, and Ruslan
  Salakhutdinov.
\newblock Dropout: a simple way to prevent neural networks from overfitting.
\newblock \emph{The Journal of Machine Learning Research}, 15\penalty0
  (1):\penalty0 1929--1958, 2014.

\bibitem[Sutskever et~al.(2013)Sutskever, Martens, Dahl, and
  Hinton]{sutskever2013importance}
Ilya Sutskever, James Martens, George Dahl, and Geoffrey Hinton.
\newblock On the importance of initialization and momentum in deep learning.
\newblock In \emph{International Conference on Machine Learning}, pp.\
  1139--1147, 2013.

\end{thebibliography}



\appendixpage
\appendix

\section{Additional details on our experiments and results}

We report the mean and the standard deviation for various performance metrics such as loss and accuracy across three random  runs of our experiments.  The hyperparameter $\eta$ is tuned for each variant of the optimizer, and the best values found are listed in \hbox{Table \ref{tab:best-etas}}.

\begin{table}[h]
\caption{\label{tab:best-etas} Best hyperparameters $\eta$ obtained for AdamUCB, AdamCB and AdamS algorithms when training MNIST and CIFAR-10 on various architectures.}
\vspace{-0.03in}
\small
\begin{center}
\small\addtolength{\tabcolsep}{-3pt}
\scalebox{0.8}{\begin{tabular}{c c c c c c c}
\toprule
\multicolumn{1}{c}{\multirow{1}{*}{\textbf{Dataset}}} &
\multicolumn{1}{c}{\multirow{1}{*}{\textbf{Model}}} &
\multicolumn{1}{c}{\multirow{1}{*}{\textbf{Batch Size}}} & \multicolumn{1}{c}{\multirow{1}{*}{\textbf{Dropout}}} & \multicolumn{1}{c}{\multirow{1}{*}{ {\boldmath{$\eta_{UCB}$}}}}    & \multicolumn{1}{c}{\boldmath{$\eta_{CB}$}}    & \multicolumn{1}{c}{\boldmath{$\eta_{S}$}}   \\ \cmidrule{1-7}
\multirow{1}{*}{{MNIST}} & \multirow{1}{*}{{LR}} & 128 & NO & $0.01$ & $0.001$ &  $0.0001$\\ \cmidrule{1-7} 
\multirow{4}{*}{{MNIST}} & \multirow{4}{*}{{MLP}} & \multirow{2.5}{*}{{128}} & NO & $0.1$ & $0.001$ &  $0.005$\\ \cmidrule{4-7}
 &&& YES & $0.1$ & $0.0005$ &  $0.005$\\ \cmidrule{3-7}
&& 16 & NO & $0.3$ & $0.0001$ &  $0.05$\\ \cmidrule{1-7}
\multirow{4}{*}{{CIFAR-10}} & \multirow{4}{*}{{CNN}} & \multirow{2.5}{*}{{128}} & NO & $0.01$ & $5\times 10^{-5}$ &  $0.0001$\\ \cmidrule{4-7}
 &&& YES & $0.05$ & $0.0001$ &  $0.0001$\\ \cmidrule{3-7}
&& 16 & NO & $0.3$ & $1\times 10^{-5}$ &  $0.005$\\ 
\bottomrule
\end{tabular}}
\end{center}
\vspace{-0.03in}
\end{table}

\subsection{Experiment: Logistic Regression}

Table \ref{tab:mnistlr-b128}  summarizes the training and validation scores at different stages of the training under the four optimizers (Adam, AdamUCB, AdamCB and AdamS) for the best values of $\eta$ given in Table \ref{tab:best-etas}. 

\begin{table}[H]
\caption{\label{tab:mnistlr-b128} Performance of our optimizers (with the best $\eta$s according to Table \ref{tab:best-etas}) compared to Adam when training a LR model on MNIST with batch size 128. The values represent $mean\pm std$ computed over three random seeded runs of the training.}
\vspace{-0.03in}
\small
\begin{center}
\small\addtolength{\tabcolsep}{-3pt}
\scalebox{0.8}{\begin{tabular}{c l c c c c} %
\toprule
 \multicolumn{1}{c}{\multirow{1}{*}{\textbf{Epoch}}} & \multicolumn{1}{c}{\multirow{1}{*}{\textbf{Optimizer}}}    & \multicolumn{1}{c}{\textbf{Train Loss}}    & \multicolumn{1}{c}{\textbf{Val. Loss}} & \multicolumn{1}{c}{\textbf{Val. Acc. (\%)}}  \\ \cmidrule{1-5}
\multirow{5}{*}{3} & Adam& ${ \boldsymbol{0.284 \pm 0.000} }$ &  ${ \boldsymbol{0.279 \pm 0.003} }$ &  ${ \boldsymbol{91.983 \pm 0.172} }$\\ \cmidrule{2-5}
 & AdamUCB &  ${ 0.298 \pm 0.001 }$ &  ${ 0.282 \pm 0.003 }$ &  ${ 91.927 \pm 0.099 }$\\ \cmidrule{2-5}
 & AdamCB &  ${ 0.333 \pm 0.004 }$ &  ${ 0.307 \pm {0.002} }$ &  ${ 91.467 \pm 0.115 }$  \\ \cmidrule{2-5}
 & AdamS &  ${ 0.298 \pm 0.001 }$ &  ${ 0.282 \pm 0.003 }$ &  ${ 91.940 \pm {0.090} }$ \\ \cmidrule{1-5}
\multirow{5}{*}{{20}} & Adam & ${ \boldsymbol{0.249 \pm 0.000} }$ &  ${ 0.273 \pm 0.003 }$ &  ${ 92.550 \pm 0.149 }$\\ \cmidrule{2-5}
 & AdamUCB &  ${ 0.250 \pm 0.001 }$ &  ${ 0.271 \pm 0.003 }$ &  ${ 92.553 \pm 0.136 }$ \\ \cmidrule{2-5}
& AdamCB &  ${ 0.255 \pm 0.001 }$ &  ${ \boldsymbol{0.270 \pm 0.003} }$ &  ${ 92.447 \pm {0.106} }$  \\ \cmidrule{2-5}
& AdamS &  ${ 0.250 \pm 0.001 }$ &  ${ 0.271 \pm 0.003 }$ &  ${ \boldsymbol{92.557 \pm 0.142} }$ \\ \cmidrule{1-5}
\multirow{5}{*}{45} & Adam& ${ \boldsymbol{0.241 \pm 0.001} }$ &  ${ 0.279 \pm 0.001 }$ &  ${ 92.500 \pm 0.114 }$\\ \cmidrule{2-5}
 & AdamUCB &  $\boldsymbol{ 0.241 \pm 0.001 }$ &  $\boldsymbol{ 0.278 \pm 0.002 }$ &  ${ 92.493 \pm 0.105 }$\\ \cmidrule{2-5}
 & AdamCB &  ${ 0.242 \pm 0.002 }$ &  ${ \boldsymbol{0.278 \pm 0.001} }$ &  ${ 92.393 \pm {0.086} }$\\ \cmidrule{2-5}
 & AdamS &  $\boldsymbol{ 0.241 \pm 0.001 }$ &  $\boldsymbol{ 0.278 \pm 0.002 }$ &  ${ \boldsymbol{92.510 \pm 0.131} }$ \\ 
\bottomrule
\end{tabular}}
\end{center}
\vspace{-0.03in}
\end{table}

\subsection{Experiment: Multi-layer Neural Networks}
We additionally experiment by adding a dropout noise  layer ($p_{drop}=0.5$) over the output activations of the first hidden layer, and study the effect of different batch sizes. 

\begin{table}[h]
\caption{\label{tab:mnist-mlp} Performance of our optimizers (with the best $\eta$s according to Table \ref{tab:best-etas}) compared to Adam when training a MLP on MNIST  under different configurations.  The values represent $mean\pm std$ computed over three random seeded runs of the training. }
\vspace{-0.03in}
\small
\begin{center}
\small\addtolength{\tabcolsep}{-3pt}
\scalebox{0.8}{\begin{tabular}{c c c l c c c c} %
\toprule
\multicolumn{1}{c}{\multirow{1}{*}{\textbf{Batch Size}}} & \multicolumn{1}{c}{\multirow{1}{*}{\textbf{Dropout}}} & \multicolumn{1}{c}{\multirow{1}{*}{\textbf{Epoch}}} & \multicolumn{1}{c}{\multirow{1}{*}{\textbf{Optimizer}}}    & \multicolumn{1}{c}{\textbf{Train Loss}}    & \multicolumn{1}{c}{\textbf{Val. Loss}} & \multicolumn{1}{c}{\textbf{Val. Acc. (\%)}}  \\ \cmidrule{1-7}
&&\multirow{5}{*}{3} & Adam& ${ 0.062 \pm 0.001 }$ &  ${ 0.077 \pm 0.008 }$ &  $\boldsymbol{ 97.667 \pm 0.178 }$\\ \cmidrule{4-7}
&& & AdamUCB &  $\boldsymbol{ 0.055 \pm 0.004 }$ &  ${ 0.075 \pm 0.006 }$ &  ${ 97.573 \pm 0.204 }$ \\ \cmidrule{4-7}
&& & AdamCB &  ${ 0.084 \pm 0.002 }$ &  ${ 0.089 \pm 0.002 }$ &  ${ 97.127 \pm 0.075 }$  \\ \cmidrule{4-7}
&& & AdamS &  ${ 0.059 \pm 0.002 }$ &  $\boldsymbol{ 0.074 \pm 0.003 }$ &  ${ 97.607 \pm 0.160 }$ \\ \cmidrule{3-7}
&&\multirow{5}{*}{{20}} & Adam & ${ 0.015 \pm 0.001 }$ &  ${ 0.077 \pm 0.011 }$ &  ${ 98.037 \pm 0.156 }$\\ \cmidrule{4-7}
&\multirow{3}{*}{\textbf{NO}} & & AdamUCB &  ${ 0.016 \pm 0.006 }$ &  $\boldsymbol{ 0.061 \pm 0.004 }$ &  ${ 98.150 \pm 0.089 }$ \\ \cmidrule{4-7}
&&& AdamCB &  ${ 0.120 \pm 0.095 }$ &  ${ 0.109 \pm 0.046 }$ &  ${ 96.663 \pm 1.425 }$ \\ \cmidrule{4-7}
&&& AdamS &  $\boldsymbol{ 0.007 \pm 0.002 }$ &  $\boldsymbol{ 0.061 \pm 0.007 }$ &  $\boldsymbol{ 98.343 \pm 0.139 }$ \\ \cmidrule{3-7}
&&\multirow{5}{*}{45} & Adam& ${ 0.009 \pm 0.003 }$ &  ${ 0.076 \pm 0.003 }$ &  ${ 98.123 \pm 0.202 }$\\ \cmidrule{4-7}
&& & AdamUCB &  ${ 0.004 \pm 0.001 }$ &  $\boldsymbol{ 0.057 \pm 0.001 }$ &  $\boldsymbol{ 98.417 \pm 0.095 }$\\ \cmidrule{4-7}
&& & AdamCB &  ${ 0.017 \pm 0.005 }$ &  ${ 0.066 \pm 0.005 }$ &  ${ 98.053 \pm 0.085 }$ \\ \cmidrule{4-7}
\multirow{3}{*}{\textbf{128}} && & AdamS &  $\boldsymbol{ 0.003 \pm 0.001 }$ &  ${ 0.063 \pm 0.004 }$ &  ${ 98.353 \pm 0.038 }$  \\ \cmidrule{2-7}

&&\multirow{5}{*}{3} & Adam& ${ 0.112 \pm 0.001 }$ &  $\boldsymbol{ 0.082 \pm 0.001 }$ &  $\boldsymbol{ 97.453 \pm 0.045 }$\\ \cmidrule{4-7}
&& & AdamUCB &  $\boldsymbol{ 0.108 \pm 0.003 }$ &  ${ 0.083 \pm 0.004 }$ &  ${ 97.333 \pm 0.125 }$ \\ \cmidrule{4-7}
&& & AdamCB &  ${ 0.123 \pm 0.004 }$ &  ${ 0.095 \pm 0.003 }$ &  ${ 97.017 \pm 0.057 }$  \\ \cmidrule{4-7}
&& & AdamS &  $\boldsymbol{ 0.108 \pm 0.003 }$ &  $\boldsymbol{ 0.082 \pm 0.006 }$ &  ${ 97.403 \pm 0.222 }$ \\ \cmidrule{3-7}
&&\multirow{5}{*}{{20}} & Adam & $\boldsymbol{ 0.054 \pm 0.003 }$ &  $\boldsymbol{ 0.060 \pm 0.001 }$ &  ${ 98.197 \pm 0.040 }$\\ \cmidrule{4-7}
&\multirow{3}{*}{\textbf{YES}} & & AdamUCB &  ${ 0.065 \pm 0.009 }$ &  $\boldsymbol{ 0.060 \pm 0.001 }$ &  $\boldsymbol{ 98.223 \pm 0.133 }$ \\ \cmidrule{4-7}
&&& AdamCB &  ${ 0.107 \pm 0.050 }$ &  ${ 0.068 \pm 0.009 }$ &  ${ 98.003 \pm 0.237 }$ \\ \cmidrule{4-7}
&&& AdamS &  ${ 0.062 \pm 0.012 }$ &  ${ 0.065 \pm 0.003 }$ &  ${ 98.120 \pm 0.090 }$\\ \cmidrule{3-7}
&&\multirow{5}{*}{45} & Adam& ${ 0.046 \pm 0.002 }$ &  ${ 0.055 \pm 0.002 }$ &  ${ 98.423 \pm 0.071 }$\\ \cmidrule{4-7}
&& & AdamUCB &  ${ 0.052 \pm 0.016 }$ &  $\boldsymbol{ 0.052 \pm 0.003 }$ &  $\boldsymbol{ 98.497 \pm 0.119 }$\\ \cmidrule{4-7}
&& & AdamCB &  ${ 0.081 \pm 0.045 }$ &  ${ 0.066 \pm 0.016 }$ &  ${ 98.060 \pm 0.418 }$ \\ \cmidrule{4-7}
&& & AdamS &  $\boldsymbol{ 0.039 \pm 0.009 }$ &  ${ 0.055 \pm 0.002 }$ &  ${ 98.470 \pm 0.108 }$ \\ \cmidrule{1-7}

&&\multirow{5}{*}{3} & Adam& $\boldsymbol{ 0.090 \pm 0.002 }$ &  ${ 0.098 \pm 0.003 }$ &  ${ 97.113 \pm 0.168 }$\\ \cmidrule{4-7}
&& & AdamUCB &  ${ 0.093 \pm 0.002 }$ &  ${ 0.094 \pm 0.007 }$ &  ${ 97.157 \pm 0.220 }$\\ \cmidrule{4-7}
&& & AdamCB &  ${ 0.117 \pm 0.005 }$ &  ${ 0.109 \pm 0.002 }$ &  ${ 96.570 \pm 0.056 }$ \\ \cmidrule{4-7}
&& & AdamS &  ${ 0.095 \pm 0.002 }$ &  $\boldsymbol{ 0.084 \pm 0.007 }$ &  $\boldsymbol{ 97.493 \pm 0.233 }$ \\ \cmidrule{3-7}
&&\multirow{5}{*}{{20}} & Adam & $\boldsymbol{ 0.042 \pm 0.001 }$ &  ${ 0.107 \pm 0.004 }$ &  $\boldsymbol{ 97.373 \pm 0.131 }$\\ \cmidrule{4-7}
\multirow{3}{*}{\textbf{16}} &\multirow{3}{*}{\textbf{NO}} & & AdamUCB &  ${ 0.043 \pm 0.002 }$ &  ${ 0.108 \pm 0.022 }$ &  ${ 97.313 \pm 0.500 }$ \\ \cmidrule{4-7}
&&& AdamCB &  ${ 0.066 \pm 0.004 }$ &  $\boldsymbol{ 0.105 \pm 0.008 }$ &  ${ 96.983 \pm 0.327 }$ \\ \cmidrule{4-7}
&&& AdamS &  ${ 0.045 \pm 0.002 }$ &  ${ 0.128 \pm 0.011 }$ &  ${ 96.910 \pm 0.456 }$\\ \cmidrule{3-7}
&&\multirow{5}{*}{45} & Adam& $\boldsymbol{ 0.035 \pm 0.000 }$ &  ${ 0.105 \pm 0.020 }$ &  $\boldsymbol{ 97.793 \pm 0.246 }$\\ \cmidrule{4-7}
&& & AdamUCB &  ${ 0.037 \pm 0.003 }$ &  ${ 0.104 \pm 0.021 }$ &  ${ 97.690 \pm 0.270 }$\\ \cmidrule{4-7}
&& & AdamCB &  ${ 0.050 \pm 0.005 }$ &  ${ 0.111 \pm 0.030 }$ &  ${ 97.117 \pm 0.692 }$\\ \cmidrule{4-7}
 && & AdamS &  ${ 0.039 \pm 0.003 }$ &  $\boldsymbol{ 0.099 \pm 0.011 }$ &  ${ 97.757 \pm 0.159 }$\\
\bottomrule
\end{tabular}}
\end{center}
\vspace{-0.03in}
\end{table}

\begin{figure}[h]
\centering

 \makebox[\textwidth][c]{
 \subfloat[Training/Validation loss and Validation accuracy when the MLP is trained with dropout at \hbox{batch size $=128$}.]{ \includegraphics[scale=0.35]{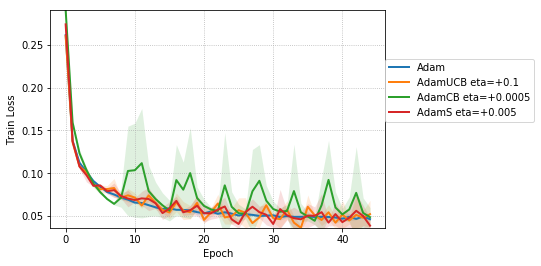}
 \includegraphics[scale=0.35]{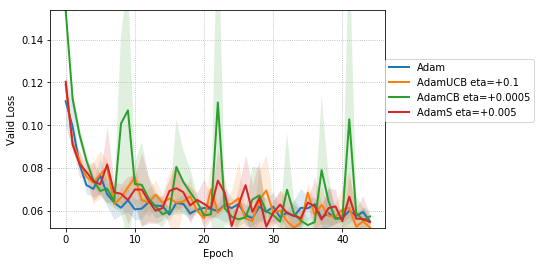}
 \includegraphics[scale=0.35]{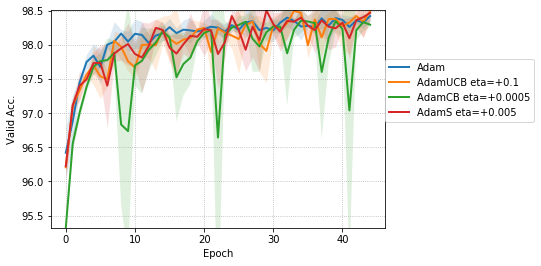}
 } 
 } \\
 
  \makebox[\textwidth][c]{
 \subfloat[Training/Validation loss and Validation accuracy when the MLP is trained without dropout at \hbox{batch size $=16$}.]{ \includegraphics[scale=0.35]{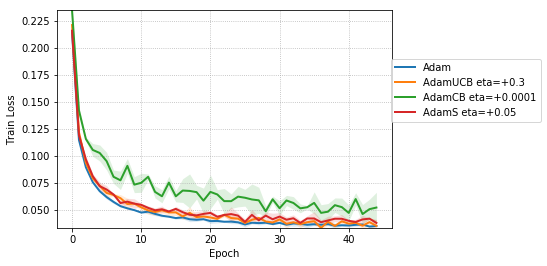}
 \includegraphics[scale=0.35]{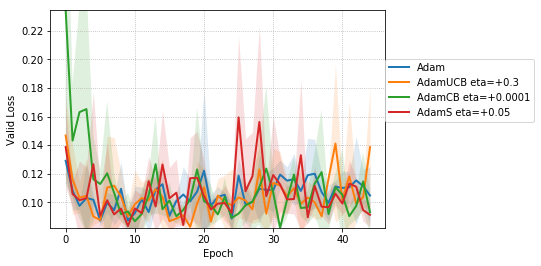}
 \includegraphics[scale=0.35]{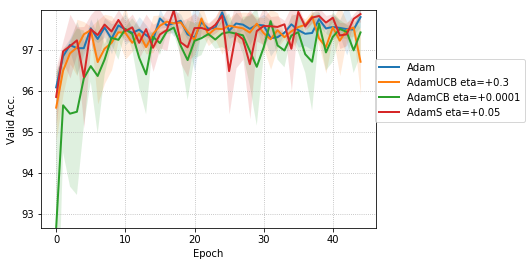}
 } 
 } 

  \caption{\label{fig:bestetasmnistmlp} Comparison of optimizers for the MLP model trained on MNIST with -- (a) batch size=128, with dropout, (b) batch size=16, without dropout.}

\end{figure}

Table \ref{tab:mnist-mlp}  and Figure \ref{fig:bestetasmnistmlp} summarize the training, and validation performances for batch sizes 16 and 128 for the best values of $\eta$ given in Table \ref{tab:best-etas} at different stages of the training.

\subsection{Experiment: Convolutional Neural Networks}

\begin{table}[hbt]
\caption{\label{tab:cifar10} Performance of our optimizers (with the best $\eta$s according to Table \ref{tab:best-etas}) compared to Adam when training a c64-c64-c128-1000 CNN on CIFAR-10  under different configurations.  The values represent $mean\pm std$ computed over three random seeded runs of the training.}
\vspace{-0.03in}
\small
\begin{center}
\small\addtolength{\tabcolsep}{-3pt}
\scalebox{0.8}{\begin{tabular}{c c c l c c c c} %
\toprule
\multicolumn{1}{c}{\multirow{1}{*}{\textbf{Batch Size}}} & \multicolumn{1}{c}{\multirow{1}{*}{\textbf{Dropout}}} & \multicolumn{1}{c}{\multirow{1}{*}{\textbf{Epoch}}} & \multicolumn{1}{c}{\multirow{1}{*}{\textbf{Optimizer}}}    & \multicolumn{1}{c}{\textbf{Train Loss}}    & \multicolumn{1}{c}{\textbf{Val. Loss}} & \multicolumn{1}{c}{\textbf{Val. Acc. (\%)}}  \\ \cmidrule{1-7}
&&\multirow{5}{*}{3} & Adam& ${ 1.133 \pm 0.091 }$ &  ${ 1.072 \pm 0.051 }$ &  ${ 62.030 \pm 1.884 }$\\ \cmidrule{4-7}
&& & AdamUCB &  ${ 1.055 \pm {0.005} }$ &  ${ 0.971 \pm 0.028 }$ &  ${ 65.800 \pm {1.025} }$ \\ \cmidrule{4-7}
&& & AdamCB &  ${ 1.000 \pm 0.013 }$ &  ${ 0.912 \pm 0.049 }$ &  ${ 67.803 \pm 2.104 }$  \\ \cmidrule{4-7}
&& & AdamS &  ${ \boldsymbol{0.994 \pm 0.015} }$ &  ${ \boldsymbol{0.903 \pm 0.024} }$ &  ${ \boldsymbol{68.073 \pm 1.395} }$ \\ \cmidrule{3-7}
&&\multirow{5}{*}{{20}} & Adam & ${ 0.646 \pm 0.078 }$ &  ${ 0.698 \pm 0.054 }$ &  ${ 76.253 \pm 1.975 }$\\ \cmidrule{4-7}
&\multirow{3}{*}{\textbf{NO}} & & AdamUCB &  ${ 0.561 \pm {0.004} }$ &  ${ 0.655 \pm 0.025 }$ &  ${ 77.843 \pm 0.939 }$ \\ \cmidrule{4-7}
&&& AdamCB &  $\boldsymbol{ 0.517 \pm {0.004} }$ &  ${ 0.646 \pm 0.012 }$ &  ${ 78.370 \pm {0.210} }$ \\ \cmidrule{4-7}
&&& AdamS &  ${ \boldsymbol{0.517 \pm 0.006} }$ &  ${ \boldsymbol{0.616 \pm 0.009} }$ &  ${ \boldsymbol{79.220 \pm 0.448} }$ \\ \cmidrule{3-7}
&&\multirow{5}{*}{45} & Adam& ${ 0.514 \pm 0.072 }$ &  ${ 0.651 \pm 0.062 }$ &  ${ 78.573 \pm 2.584 }$\\ \cmidrule{4-7}
&& & AdamUCB &  ${ 0.425 \pm 0.009 }$ &  ${ 0.589 \pm {0.012} }$ &  ${ 81.073 \pm {0.505} }$\\ \cmidrule{4-7}
&& & AdamCB &  ${ 0.397 \pm 0.028 }$ &  ${ 0.594 \pm 0.026 }$ &  ${ 80.987 \pm 0.885 }$ \\ \cmidrule{4-7}
\multirow{3}{*}{\textbf{128}} && & AdamS &  ${ \boldsymbol{0.383 \pm 0.006} }$ &  ${ \boldsymbol{0.586 \pm 0.025} }$ &  ${ \boldsymbol{81.803 \pm 0.665} }$ \\ \cmidrule{2-7}

&&\multirow{5}{*}{3} & Adam& ${ 1.183 \pm 0.015 }$ &  ${ 1.061 \pm \boldsymbol{0.021} }$ &  ${ 62.470 \pm \boldsymbol{0.489} }$\\ \cmidrule{4-7}
&& & AdamUCB &  ${ 1.171 \pm 0.025 }$ &  ${ 1.060 \pm 0.031 }$ &  ${ 62.383 \pm 0.957 }$ \\ \cmidrule{4-7}
&& & AdamCB &  ${ \boldsymbol{1.104 \pm 0.025} }$ &  ${ \boldsymbol{1.020 \pm 0.052} }$ &  ${ \boldsymbol{64.110 \pm 1.187} }$  \\ \cmidrule{4-7}
&& & AdamS &  ${ 1.143 \pm {0.013} }$ &  ${ 1.035 \pm 0.023 }$ &  ${ 63.133 \pm 0.740 }$ \\ \cmidrule{3-7}
&&\multirow{5}{*}{{20}} & Adam & ${ 0.756 \pm 0.013 }$ &  ${ 0.709 \pm 0.022 }$ &  ${ 75.500 \pm 0.568 }$\\ \cmidrule{4-7}
&\multirow{3}{*}{\textbf{YES}} & & AdamUCB &  ${ 0.743 \pm 0.014 }$ &  ${ 0.686 \pm 0.019 }$ &  ${ 76.430 \pm 0.661 }$ \\ \cmidrule{4-7}
&&& AdamCB &  ${ \boldsymbol{0.677 \pm 0.006} }$ &  ${ \boldsymbol{0.641 \pm 0.011} }$ &  ${ \boldsymbol{78.063 \pm 0.422} }$ \\ \cmidrule{4-7}
&&& AdamS &  ${ 0.699 \pm 0.039 }$ &  ${ 0.653 \pm 0.036 }$ &  ${ 77.727 \pm 1.279 }$ \\ \cmidrule{3-7}
&&\multirow{5}{*}{45} & Adam& ${ 0.650 \pm 0.006 }$ &  ${ 0.599 \pm 0.027 }$ &  ${ 79.617 \pm 0.950 }$\\ \cmidrule{4-7}
&& & AdamUCB &  ${ 0.639 \pm 0.008 }$ &  ${ 0.595 \pm {0.008} }$ &  ${ 79.947 \pm {0.240} }$\\ \cmidrule{4-7}
&& & AdamCB &  ${ \boldsymbol{0.573 \pm 0.001} }$ &  ${ \boldsymbol{0.549 \pm 0.014} }$ &  ${ \boldsymbol{81.203 \pm 0.565} }$ \\ \cmidrule{4-7}
&& & AdamS &  ${ 0.593 \pm 0.032 }$ &  ${ 0.566 \pm 0.025 }$ &  ${ 80.793 \pm 0.926 }$ \\ \cmidrule{1-7}

&&\multirow{5}{*}{3} & Adam& ${ 1.447 \pm 0.059 }$ &  ${ 1.357 \pm 0.064 }$ &  ${ 50.920 \pm 3.231 }$\\ \cmidrule{4-7}
&& & AdamUCB &  ${ 1.390 \pm 0.042 }$ &  ${ 1.283 \pm 0.057 }$ &  ${ 54.137 \pm 2.479 }$ \\ \cmidrule{4-7}
&& & AdamCB &  ${ 1.335 \pm {0.019} }$ &  ${ 1.236 \pm {0.014} }$ &  ${ 56.003 \pm {1.024} }$ \\ \cmidrule{4-7}
&& & AdamS &  ${ \boldsymbol{1.319 \pm 0.031} }$ &  ${ \boldsymbol{1.225 \pm 0.052} }$ &  ${ \boldsymbol{56.763 \pm 1.792} }$ \\ \cmidrule{3-7}
&&\multirow{5}{*}{{20}} & Adam & ${ 1.050 \pm 0.073 }$ &  ${ 1.001 \pm 0.057 }$ &  ${ 65.203 \pm 1.932 }$\\ \cmidrule{4-7}
\multirow{3}{*}{\textbf{16}} &\multirow{3}{*}{\textbf{NO}} & & AdamUCB &  ${ 0.922 \pm 0.054 }$ &  ${ 0.914 \pm 0.069 }$ &  ${ 68.520 \pm 2.721 }$ \\ \cmidrule{4-7}
&&& AdamCB &  ${ 0.923 \pm 0.022 }$ &  ${ 0.918 \pm 0.018 }$ &  ${ 68.757 \pm {0.273} }$ \\ \cmidrule{4-7}
&&& AdamS &  ${ \boldsymbol{0.918 \pm 0.020} }$ &  ${ \boldsymbol{0.897 \pm 0.015} }$ &  ${ \boldsymbol{69.443 \pm 0.425} }$ \\ \cmidrule{3-7}
&&\multirow{5}{*}{45} & Adam& ${ 0.825 \pm 0.026 }$ &  ${ 0.841 \pm {0.019} }$ &  ${ 71.887 \pm {0.870} }$\\ \cmidrule{4-7}
&& & AdamUCB &  ${ 0.766 \pm 0.041 }$ &  ${ 0.784 \pm 0.040 }$ &  ${ 73.393 \pm 1.351 }$\\ \cmidrule{4-7}
&& & AdamCB &  ${ 0.783 \pm {0.009} }$ &  ${ 0.801 \pm 0.037 }$ &  ${ 72.793 \pm 1.072 }$ \\ \cmidrule{4-7}
 && & AdamS &  ${ \boldsymbol{0.765 \pm 0.015} }$ &  ${ \boldsymbol{0.768 \pm 0.033} }$ &  ${ \boldsymbol{74.063 \pm 1.095} }$\\
\bottomrule
\end{tabular}}
\end{center}
\vspace{-0.03in}
\end{table}

\begin{figure}[h]
\centering
 
  \makebox[\textwidth][c]{
 \subfloat[Training loss]{ \includegraphics[scale=0.35]{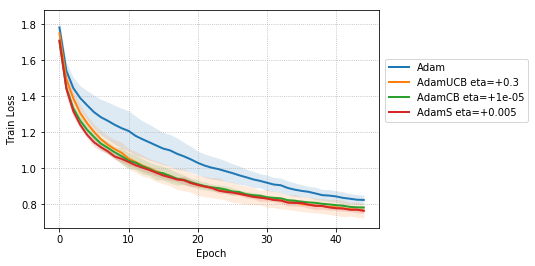}}
 \subfloat[Validation loss]{\includegraphics[scale=0.35]{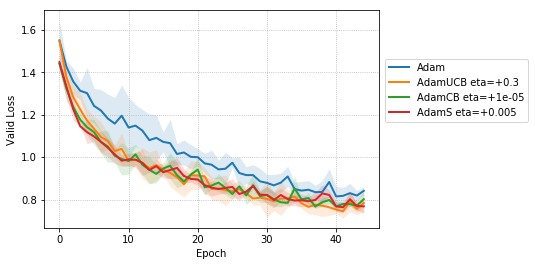}}
 \subfloat[Validation acc.]{\includegraphics[scale=0.35]{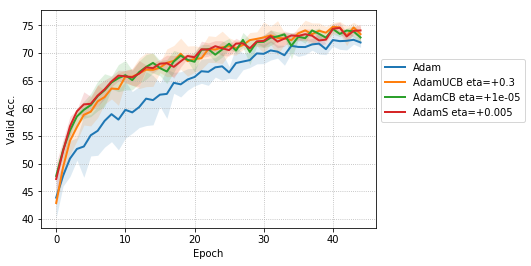}
 } 
 } 

  \caption{\label{fig:bestetasperformancecifar10} Training/Validation loss and Validation accuracy when the CNN is trained without dropout at \hbox{batch size $=16$}.}
  
\end{figure}

\begin{figure}[tp]
\centering
\makebox[\textwidth][c]{
 \subfloat[Performance of AdamUCB optimizer across different $\eta$ compared to Adam.]{ \includegraphics[scale=0.21]{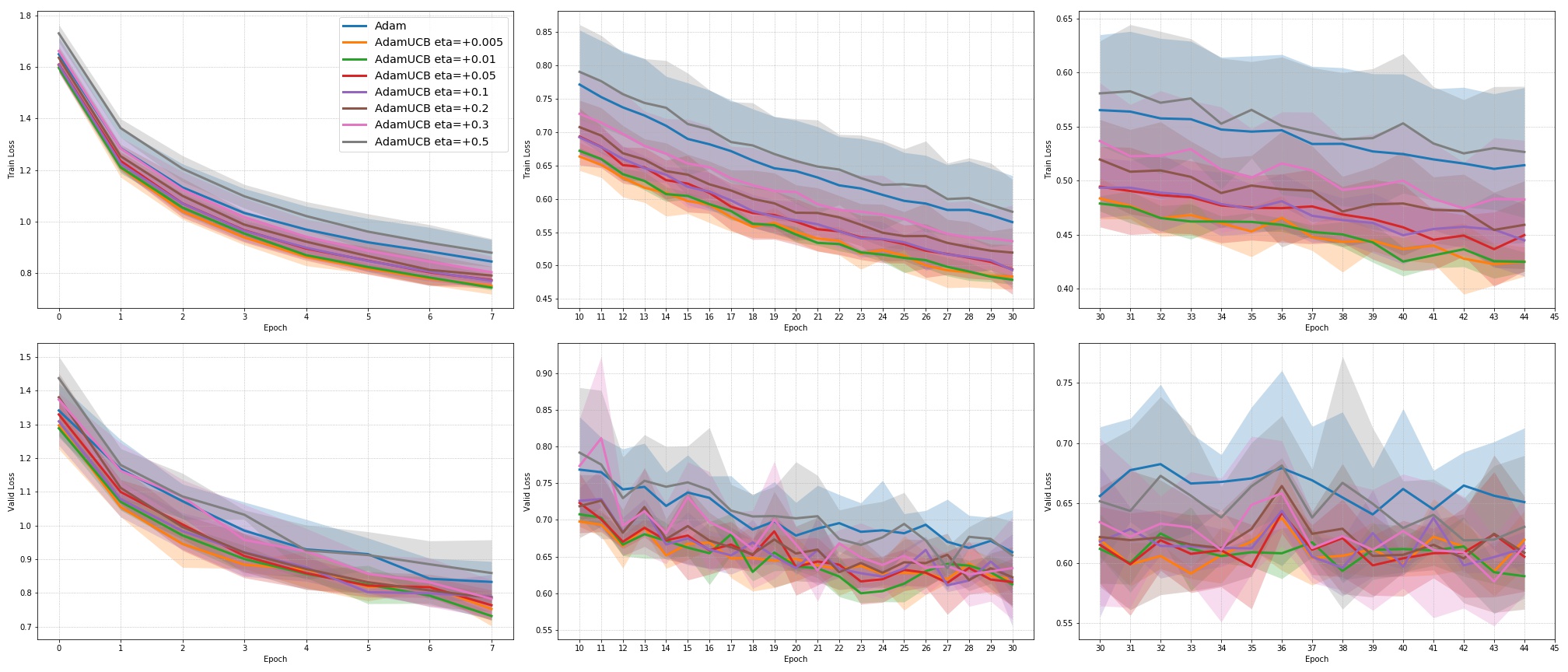}} } \\
 
 \makebox[\textwidth][c]{
  \subfloat[Performance of AdamCB optimizer across different $\eta$ compared to Adam.]{ \includegraphics[scale=0.21]{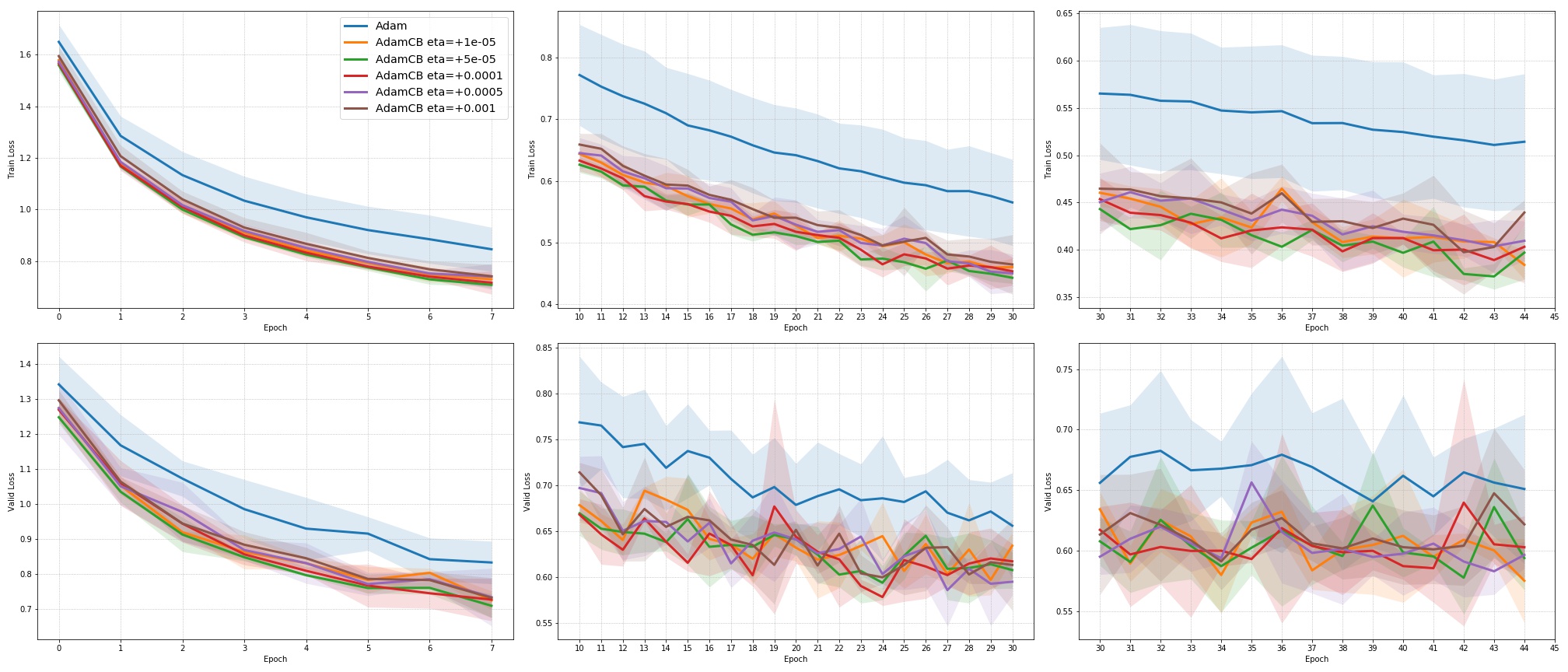}}} \\
  
  \makebox[\textwidth][c]{
  \subfloat[Performance of AdamS optimizer across different $\eta$ compared to Adam.]{ \includegraphics[scale=0.21]{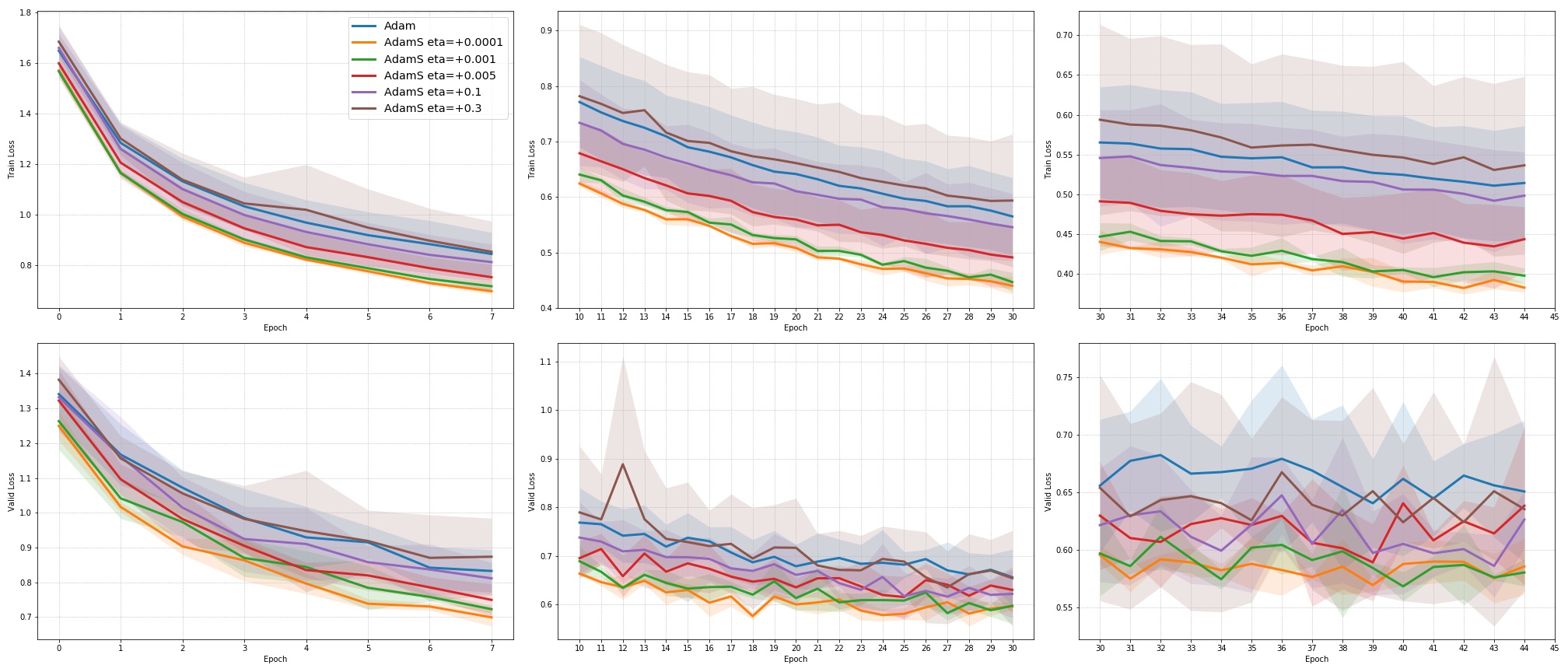}}} 

  \caption{Training of  c64-c64-c128-1000 CNN architecture (without dropout) on CIFAR-10 images with \hbox{batch size $=128$}. Performance of  (a) \textit{AdamUCB}, (b) \textit{AdamCB} and (c) \textit{AdamS} algorithms  compared to the original \textit{Adam}  optimizer across training and validation.}
  
  \label{fig:allmodels}
\end{figure}

We  include results of our experiments with batch size = 16. Figure \ref{fig:bestetasperformancecifar10} shows the improvement in the rate of convergence for the case of batch size = 16. Figure \ref{fig:allmodels} summarizes the results of our experiments without dropout on a mini-batch size of $128$ across different $\eta$. Table \ref{tab:cifar10}   summarize the training, and validation performances for batch sizes 16 and 128 for the best values of $\eta$ given in Table \ref{tab:best-etas} at different stages of the training. 

\section{Discussion}

For the case of MLPs with dropout, AdamS  outperforms Adam but with a reduced margin of improvement (see Table \ref{tab:mnist-mlp}) when compared to the case without dropout. When a batch size of 16 is used instead, AdamS and Adam perform fairly similar.

For the case of CNNs  trained on CIFAR-10, clearly, from the Figure \ref{fig:allmodels}, AdamUCB, AdamCB and AdamS perform significantly better for appropriate $\eta$s than the original Adam optimizer. Figure \ref{fig:bestetasperformancecifar10} shows how the proposed versions of Adam accelerate training for batch size = 16. We also notice that the improvement in performance for lower mini-batch sizes is more significant for CIFAR-10.

Table \ref{tab:cifar10} clearly shows how our variants of Adam accelerate training, especially, in the initial few epochs for CIFAR-10 dataset for both the cases of mini-batch sizes 16 and 128. When dropout is used, the improvement in validation accuracy reduces by some margin. \qedsymbol

\end{document}